\pdfoutput=1

\documentclass[11pt]{article}
\usepackage{graphicx}
\usepackage{float}
\usepackage[final]{acl}

\usepackage{times}
\usepackage{latexsym}
\usepackage{hyperref}

\usepackage[T1]{fontenc}

\usepackage[utf8]{inputenc}

\usepackage{microtype}

    \title{Leverage Points in Modality Shifts: \\ Comparing Language-only and Multimodal Word Representations}

\author{
  Aleksey Tikhonov \\
  Inworld.AI \\
  \texttt{altsoph@gmail.com} \\\AND
Lisa Bylinina\thanks{ \ \ Equal contribution.} \\
 Center for Language and Cognition \\University of Groningen \\
  \texttt{e.g.bylinina@rug.nl} \\\And
  Denis Paperno$ ^\ast$ \\
  Institute for Language Sciences \\ Utrecht University \\
  \texttt{d.paperno@uu.nl} \\
  }

\begin{document}
\maketitle
\begin{abstract}
Multimodal embeddings aim to enrich the semantic information in neural representations of language compared to text-only models. While different embeddings exhibit different applicability and performance on downstream tasks, little is known about the systematic representation differences attributed to the visual modality.  Our paper compares word embeddings from three vision-and-language models (CLIP, OpenCLIP and Multilingual CLIP, \citealt{radford2021learning,ilharco_gabriel_2021_5143773,carlsson-etal-2022-cross}) and three text-only models, with static (FastText, \citealp{bojanowski2017enriching}) as well as contextual representations (multilingual BERT \citealt{mbert}; XLM-RoBERTa, \citealt{xlmroberta}). This is the first large-scale study of the effect of visual grounding on language representations, including 46 semantic parameters. We identify meaning properties and relations that characterize words whose embeddings are most affected by the inclusion of visual modality in the training data; that is, points where visual grounding turns out most important. We find that the effect of visual modality correlates most with denotational semantic properties related to concreteness, but is also detected for several specific semantic classes, as well as for valence, a sentiment-related connotational property of linguistic expressions.
\end{abstract}

\section{Introduction}

Linguistic representations developed by recent large pre-trained language models (LMs) (\citealp{mbert,roberta,gpt} a.o.) proved to be very useful across a variety of practical applications. This success has given a new life to the debate around extractability and quality of semantic information in representations trained solely on textual input. According to the widely supported argument,   unless the textual data is grounded in a separate space (say, visual), the linguistic representations are bound to be semantically deficient (see \citealp{climbing} a.o.). 

We aim to shed new empirical light on the discussion of grounding in computational models by comparing language-only text representations to visually informed text representations. Recent advances produced empirically successful large models pre-trained on a combination of textual and visual data \citep{li2019visualbert,tan2019lxmert,tan2020vokenization,radford2021learning}. While these multimodal systems have already given rise to a plethora of applications for language-and-vision (L\&V) downstream tasks, there is still little work that directly compares textual representations of language-only models to those of multimodal ones (however, see  \citealp{davis2019deconstructing,LUDDECKE201944,10.1162/tacl_a_00443}). 
In contrast to previous related work that focuses on model evaluation with respect to specific benchmarks, we look at the impact of visual grounding from a somewhat different, non-evaluation-based perspective. We do not aim to measure the representation quality 
with respect to some gold standard, but compare language-only and L\&V models to each other intrinsically. 
Our \textbf{goal} is to identify the areas in which \textit{the contrasts between the two kinds of models} tend to lie, independent of the models' fitness for specific tasks.

To do so, we focus on a set of 13k word pairs and compare cosine distances within these pairs in the embedding spaces of language-only vs.\ L\&V models. Fixing the word pairs and comparing the models allows us to measure how the change in model modality stretches the embedding space, with the word pairs as indirect reference points. 

The pairs are characterized along 46 different semantic parameters. This information makes it possible to identify the meaning aspects for which the change in model modality matters the most. 

Our \textbf{contributions} are: 
\begin{enumerate}
\item a methodology for measuring the influence of grounding on semantic representations; 
\item a dataset characterizing a large number of word pairs along various semantic parameters and embedding distances in the models that we study.
\end{enumerate}
Our {\bf results} are the following: 

\vspace*{1ex}
\noindent $\bullet$  
The semantic parameter that makes the highest contribution into explaining the impact of modality on word representation is {\bf concreteness}. This aligns with previous results that visual modality improves representations of concrete nouns but not abstract ones \citep{10.1162/tacl_a_00443}.

\vspace*{1ex}
\noindent $\bullet$  Representations of particular semantic groups of nouns are affected the most. 

\vspace*{1ex}
\noindent $\bullet$  Semantic relations between nouns  only have small interaction with modality across the models we tested, with variation from model to model.

\vspace*{1ex}
\noindent $\bullet$ Connotational meanings from the 
VAD (valence, arousal, dominance) repertoire \citep{vad} -- specifically, valence --  play a role in representational shifts relating to modality. This is a somewhat surprising result since visual grounding is expected to relate to the denotational aspects of representations. This result is in line with recent discussion in semantics about the inter-relatedness of denotational and connotational meanings \citep{ruytenbeek2017asymmetric,terkourafi2020different,van2021adjectival,beltrama2021just,gotzner2021face}.

\vspace*{1ex}
\noindent We now discuss our data, analysis and results.

\section{Data\footnote{Our code and data are available on GitHub: \url{https://github.com/altsoph/modality_shifts}}}

The dataset consists of word pairs. To collect them, we start with 1000 most frequent words in FastText \citep{bojanowski2017enriching}. For each of them, we take 100 closest words, by cosine distance over FastText embeddings. This gives 1M pairs to work with. We filter this list of pairs in several ways. First, we only keep those pairs where both words are nouns, according to both NLTK\footnote{\url{https://github.com/nltk/nltk}} and SpaCy\footnote{\url{https://github.com/explosion/spacy-models}} POS labels. 
Second, we filter out pairs where one of the words is a substring of the other or where the two words have the same lemma. This helps against some FastText artifacts. 

One of the goals of our filtering strategy was to balance representation quality of the words (the frequency filter) and the chance for the pair to stand in a WordNet relation (the similarity filter). This gives us a set of pairs like the following:

\begin{center}
   $\langle$ page,	article $\rangle$ \\
$\langle$ people,	politicians $\rangle$ \\
$\langle$ city,	hometown $\rangle$
\end{center}

Each of the resulting pairs was characterized along a set of properties of interest, collected over a variety of available sources of human-annotated semantic information. The properties we look at come in two big blocks: 1) the ones that characterize individual words (assigned to each word in the pair); 2) the ones that characterize a semantic relation between the words in the pair. 

Properties for individual words included:

\vspace*{1ex}
\noindent $\bullet$  {\bf Concreteness}, a continuous score on the abstractness-concreteness scale, the Ghent concreteness norms \citep{concreteness};

\vspace*{1ex}
\noindent $\bullet$ 26 {\bf WordNet supersenses} of nouns ({\sc act}, {\sc animal}, {\sc feeling}, {\sc food} etc.), implemented as boolean labels \citep{wordnet};

\vspace*{1ex}
\noindent $\bullet$  3 NRC VAD continuous scores for {\bf valence}, {\bf arousal} and {\bf dominance} \citep{vad}.

\vspace*{1ex}
\noindent Relational semantic properties included:

\vspace*{1ex}
\noindent $\bullet$  6 {\bf WordNet semantic relations} \citep{wordnet}: {\sc antonyms, synonyms, same\_hyponyms, same\_hypernyms, hyponyms, hypernyms}.

\vspace*{1ex}
\noindent $\bullet$  10 {\bf ConceptNet semantic relations} \citep{conceptnet}: {\sc Antonym, Synonym, AtLocation, DerivedFrom, DistinctFrom, FormOf, IsA, PartOf, RelatedTo, SimilarTo}.

\vspace*{1ex}
\noindent The relations were implemented as boolean labels.

This is the most comprehensive list of semantic parameters  for which human annotations exist on a large scale. It covers both denotational and connotational aspects of meaning of both individual words and relation within pairs. Connotational meanings are represented with three sentiment-related meaning aspects only, as these are the only ones represented in a large human-annotated dataset \citep{vad}.

Additionally, word count based on Wikipedia (accessed via {\tt Textacy}) is included for each word in all pairs as a non-semantic baseline parameter.

We leave only those word pairs for which all the above mentioned parameters are defined. This gives us 13k word pairs in total, each of the pairs gets characterized along 30 individual semantic parameters (*2, for the first and the second noun in the pair) and 16 relational parameters; plus, word count for each of the words in the pair. 

We collect the distances between the words in each pair for their embeddings from the models of interest. 
As \textbf{text-only models}, we use fastText \citep{bojanowski2017enriching} and two contextualized embedding models: multilingual BERT (mBERT, \citealp{mbert}) and XLM-RoBERTa (XLMR, \citealp{xlmroberta}). For each contextualized model, we extract three kinds of word type embeddings known to show systematic differences \citep{vulic2020probing}; 
average of all token embeddings, including separator tokens, from the final encoding layer of a word presented in isolation (\textbf{iso});
the average encoding over the bottom 6 layers across a sample of 10 usage contexts  
(\textbf{avg-bottom}), 
amd 
the average encoding from the final layer across a sample of 10 usage contexts
(\textbf{avg-last}).
As multimodal models, we use CLIP, OpenCLIP and Multilingual CLIP \citep{radford2021learning,ilharco_gabriel_2021_5143773,carlsson-etal-2022-cross}. For each multimodal model, we extract two different types of word type embeddings, one by encoding the word in isolation 
and one by averaging over sentence embeddings of 10 usage examples. 

The goal is to find a common ground of different models depending on their modality. 
In this way we hope to be able to distinguish between model-specific idiosyncrasies and general properties of text-based representations. 

\section{Analysis}

We run a series of regression analyses with semantic features and relations as predictors, along with word frequency as baseline. 

We analyze the shift in distances within word pairs between two embedding models. To measure it, we rank all word pairs in our dataset by the ratio between the cosine distance values of the pair in the two embedding models. Using ratios and ranks rather than absolute differences serves as a normalization strategy because the vector spaces have significantly different structures (see Appendix A). The resulting rank of the pair is then used as the dependent variable in a regression analysis. 

The baseline regression model includes as predictors word frequencies in the Wikipedia corpus and concreteness scores from the Ghent concreteness norms dataset \citep{concreteness}. To estimate the contribution of different groups of semantic features, we add them to the regression as additional predictors. This is done separately for 
\begin{enumerate}
\item taxonomic features of the two words formalized as their WordNet supersenses \citep{wordnet}; 
\item sentiment/connotation-related features of the two words extracted from NRC VAD \citep{vad}; 
\item relation within the word pair according to Princeton WordNet \citep{wordnet}; 
\item relation within the word pair according to ConceptNet \citep{conceptnet}.
\end{enumerate}

All numeric parameters (concreteness scores, word frequencies, and VAD values) were normalized by converting numeric values into ranks. 

To calculate regression, we used a standard implementation of  ordinary least squares regression from the statsmodels python package. We compute adjusted R-squared values to avoid a bias from the different numbers of parameters. 
Each fitted regression showed high significance ($p<0.0001$).

\section{Results}

\begin{table*}
\centering
\begin{tabular}{lcccc}
\hline
\textbf{CLIP-iso} vs. & \textbf{XLMR-iso} & \textbf{mBERT-iso} & \textbf{BERT-avg-last}& \textbf{fastText}\\
\hline
Baselines \\
\hline
concreteness & 9.5  & 11.68 & 2.27  & 8.71\\ 
frequency & 5.43 &  7.81 & 1.91 & 0.45\\ 
concreteness+frequency & 16.73 & 17.16 & 3.65 & 9.54\\ 
\hline
+taxonomic & 21 (+4.27) & 20.35 (+3.19) & 5.43 (+1.78) & 19.50 (+9.96)\\ 
+VAD & 17.36 (+0.63) & 17.49 (+0.33) & 4.62 (+0.97) & 10.78 (+1.24)\\ 
+WordNet relations & 18.47 (+1.74) & 17.36 (+0.2)  & 10.05 (+6.4) & 10.34 (+0.8)\\ 
+ConceptNet relations & 19.8  (+3.07) & 17.47 (+0.31) & 8.84 (+5.19) & 10.26 (+0.72)\\
\hline
\end{tabular}
\caption{Illustration of our method: Embedding space in CLIP-iso vs.~four of the text-only models. Table reports percentage of variance (adjusted $R^2$) in cosine distance ratio 
explained by different groups of semantic factors. We take the number in parentheses as an estimate of the \textit{effect} of the factor (e.g.~the effect of all taxonomic features from WordNet combined) on the difference between two embedding spaces (e.g.~fastText vs.~CLIP).}
\label{tab:results}
\end{table*}

\begin{figure*}[t]
\includegraphics[width=.246\textwidth]{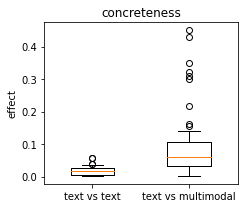} \includegraphics[width=.246\textwidth]{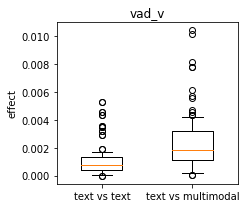}  \includegraphics[width=.246\textwidth]{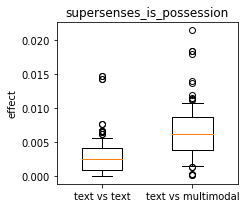} \includegraphics[width=.246\textwidth]{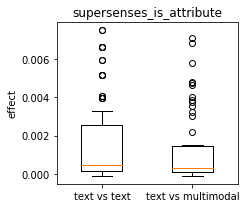}
\caption{Comparing semantic features' contributions to contrasts between text models vs. other text models, on the one hand, and text models vs. L\&V models, on the other hand. Explanatory contributions of concreteness, VAD valence and Wordnet supersense `Is Possession' are sensitive to model modality, unlike supersense `Is Attribute'. (Here and in Appendix B, whiskers in the boxplots are set to 0.5 IQR.)}
\label{fig:whiskers}
\end{figure*}

The results of regression analysis for several models are illustrated in Table \ref{tab:results}. Our main observations are:

\vspace*{1ex}
\noindent $\bullet$ {\bf Baselines}. Concreteness plays a major role in explaining modality shifts, in line with results of previous studies \citep{10.1162/tacl_a_00443}. 

\vspace*{1ex}
\noindent $\bullet$ Combined {\bf WordNet supersenses}. We find a significant effect for many pairs of text vs.~multimodal models, although different subsets of taxonomic features prove significant in different pairs of models. 

\vspace*{1ex}
\noindent $\bullet$  {\bf WordNet and ConceptNet relations} tend to be significant when aggregated, although no individual relation has a systematic effect across model pairs. 


\vspace*{.5ex}
\noindent  $\bullet$ {\bf VAD} features produce varied effects, with valence showing the most consistent modality difference. 
VAD features explain only a small percentage of variance in all models. 

Figure \ref{fig:whiskers} illustrates the effect of specific features: concreteness, valence and possession WordNet supersense, vs.\ the attribute supersense that has no consistent effect on modality shifts. For more plots, see Appendix B.

\section{Conclusion and discussion}
The goal of our paper was to investigate what semantic factors contribute to the difference in representational spaces of language-only models vs.\ multimodal models. 

Our regression analysis confirmed previous findings that concreteness plays a major role in this difference \citep{10.1162/tacl_a_00443}. This is natural since imageability, the measurable manifestation of concreteness, is directly related to whether useful information about a concept can be inferred form visual data.

However, other factors beyond abstractness contribute to the modality-based space contrasts as well. The most important factor here is taxonomical, as measured by the effect of WordNet lexicographer files.  Wordnet supersenses consistently affect semantic similarities in text-only models vs. L\&V models: in particular, we found this for artifacts, quantities, possessions and communication lexical classes.



Lastly, sentiment-related lexical properties, most clearly valence, also affect the semantic similarity in language-only vs.\ multimodal spaces. Recently, several studies in semantics and pragmatics have indicated interactions of connotational content with denotational meanings \citep{ruytenbeek2017asymmetric,terkourafi2020different,van2021adjectival,beltrama2021just,gotzner2021face}. Our results can be interpreted as pointing in that direction too. Still, the effect of sentiment is overall much smaller than the core denotational properties of the words in the lexical pair, as illustrated by the comparison of the combined VAD to combined taxonomic features in Table \ref{tab:results}. 

We contribute to the understanding of different embedding spaces  by demonstrating systematic differences between text-only vs. L\&V models. Many questions are however left for future research. For example, do the distinct properties of multimodal embeddings make them better suited for specific tasks, as \citet{10.1162/tacl_a_00443} argued for the relatedness judgments of concrete nouns?

In the light of Kruszewski's finding \citep{kruszewski2015so} that taxonomic information interacts strongly with referential compatibility between concepts, our findings on the role of taxonomic status on vector space structure suggests that the choice of multimodal vs.\ textual representations can be crucial for inference, especially for the difficult case of the neutral vs.\ contradiction distinction.

Finally, we note that the semantic factors we considered only explain a small part of the discrepancy between textual and L\&V models. The rest must be attributed to other factors, such as random differences in the textual data used for model training as well as semantic phenomena outside the scope of our study. 

We hope that our study inspires further exploration of systematic differences between embedding models, both for visual grounding and beyond.

\section*{Acknowledgements}
This work was funded by the European Research Council (ERC) under the European Union’s Horizon
2020 research and innovation programme (grant agreement No 742204). 

\bibliography{anthology,custom}

\begin{thebibliography}{26}
\expandafter\ifx\csname natexlab\endcsname\relax\def\natexlab#1{#1}\fi

\bibitem[{Beltrama(2021)}]{beltrama2021just}
Andrea Beltrama. 2021.
\newblock Just perfect, simply the best: an analysis of emphatic exclusion.
\newblock \emph{Linguistics and Philosophy}, pages 1--44.

\bibitem[{Bender and Koller(2020)}]{climbing}
Emily~M Bender and Alexander Koller. 2020.
\newblock Climbing towards nlu: On meaning, form, and understanding in the age
  of data.
\newblock In \emph{Proceedings of the 58th Annual Meeting of the Association
  for Computational Linguistics}, pages 5185--5198.

\bibitem[{Bojanowski et~al.(2017)Bojanowski, Grave, Joulin, and
  Mikolov}]{bojanowski2017enriching}
Piotr Bojanowski, Edouard Grave, Armand Joulin, and Tomas Mikolov. 2017.
\newblock Enriching word vectors with subword information.
\newblock \emph{Transactions of the Association for Computational Linguistics},
  5:135--146.

\bibitem[{Brysbaert et~al.(2014)Brysbaert, Warriner, and
  Kuperman}]{concreteness}
Marc Brysbaert, Amy~Beth Warriner, and Victor Kuperman. 2014.
\newblock Concreteness ratings for 40 thousand generally known english word
  lemmas.
\newblock \emph{Behavior research methods}, 46(3):904--911.

\bibitem[{Carlsson et~al.(2022)Carlsson, Eisen, Rekathati, and
  Sahlgren}]{carlsson-etal-2022-cross}
Fredrik Carlsson, Philipp Eisen, Faton Rekathati, and Magnus Sahlgren. 2022.
\newblock \href {https://aclanthology.org/2022.lrec-1.739} {Cross-lingual and
  multilingual {CLIP}}.
\newblock In \emph{Proceedings of the Thirteenth Language Resources and
  Evaluation Conference}, pages 6848--6854, Marseille, France. European
  Language Resources Association.

\bibitem[{Conneau et~al.(2019)Conneau, Khandelwal, Goyal, Chaudhary, Wenzek,
  Guzm{\'{a}}n, Grave, Ott, Zettlemoyer, and Stoyanov}]{xlmroberta}
Alexis Conneau, Kartikay Khandelwal, Naman Goyal, Vishrav Chaudhary, Guillaume
  Wenzek, Francisco Guzm{\'{a}}n, Edouard Grave, Myle Ott, Luke Zettlemoyer,
  and Veselin Stoyanov. 2019.
\newblock \href {http://arxiv.org/abs/1911.02116} {Unsupervised cross-lingual
  representation learning at scale}.
\newblock \emph{CoRR}, abs/1911.02116.

\bibitem[{Davis et~al.(2019)Davis, Bulat, Ver{\H{o}}, and
  Shutova}]{davis2019deconstructing}
Christopher Davis, Luana Bulat, Anita~Lilla Ver{\H{o}}, and Ekaterina Shutova.
  2019.
\newblock Deconstructing multimodality: visual properties and visual context in
  human semantic processing.
\newblock In \emph{Proceedings of the Eighth Joint Conference on Lexical and
  Computational Semantics (* SEM 2019)}, pages 118--124.

\bibitem[{Devlin et~al.(2018)Devlin, Chang, Lee, and Toutanova}]{mbert}
Jacob Devlin, Ming{-}Wei Chang, Kenton Lee, and Kristina Toutanova. 2018.
\newblock \href {http://arxiv.org/abs/1810.04805} {{BERT:} pre-training of deep
  bidirectional transformers for language understanding}.
\newblock \emph{CoRR}, abs/1810.04805.

\bibitem[{Gotzner and Mazzarella(2021)}]{gotzner2021face}
Nicole Gotzner and Diana Mazzarella. 2021.
\newblock Face management and negative strengthening: The role of power
  relations, social distance, and gender.
\newblock \emph{Frontiers in psychology}, 12.

\bibitem[{Ilharco et~al.(2021)Ilharco, Wortsman, Wightman, Gordon, Carlini,
  Taori, Dave, Shankar, Namkoong, Miller, Hajishirzi, Farhadi, and
  Schmidt}]{ilharco_gabriel_2021_5143773}
Gabriel Ilharco, Mitchell Wortsman, Ross Wightman, Cade Gordon, Nicholas
  Carlini, Rohan Taori, Achal Dave, Vaishaal Shankar, Hongseok Namkoong, John
  Miller, Hannaneh Hajishirzi, Ali Farhadi, and Ludwig Schmidt. 2021.
\newblock \href {https://doi.org/10.5281/zenodo.5143773} {Openclip}.
\newblock If you use this software, please cite it as below.

\bibitem[{Kruszewski and Baroni(2015)}]{kruszewski2015so}
Germ{\'a}n Kruszewski and Marco Baroni. 2015.
\newblock So similar and yet incompatible: Toward the automated identification
  of semantically compatible words.
\newblock In \emph{Proceedings of the 2015 Conference of the North American
  Chapter of the Association for Computational Linguistics: Human Language
  Technologies}, pages 964--969.

\bibitem[{Li et~al.(2019)Li, Yatskar, Yin, Hsieh, and Chang}]{li2019visualbert}
Liunian~Harold Li, Mark Yatskar, Da~Yin, Cho-Jui Hsieh, and Kai-Wei Chang.
  2019.
\newblock Visualbert: A simple and performant baseline for vision and language.
\newblock \emph{arXiv preprint arXiv:1908.03557}.

\bibitem[{Liu et~al.(2019)Liu, Ott, Goyal, Du, Joshi, Chen, Levy, Lewis,
  Zettlemoyer, and Stoyanov}]{roberta}
Yinhan Liu, Myle Ott, Naman Goyal, Jingfei Du, Mandar Joshi, Danqi Chen, Omer
  Levy, Mike Lewis, Luke Zettlemoyer, and Veselin Stoyanov. 2019.
\newblock Roberta: A robustly optimized bert pretraining approach.
\newblock \emph{arXiv preprint arXiv:1907.11692}.

\bibitem[{Lüddecke et~al.(2019)Lüddecke, Agostini, Fauth, Tamosiunaite, and
  Wörgötter}]{LUDDECKE201944}
Timo Lüddecke, Alejandro Agostini, Michael Fauth, Minija Tamosiunaite, and
  Florentin Wörgötter. 2019.
\newblock \href {https://doi.org/https://doi.org/10.1016/j.artint.2018.12.009}
  {Distributional semantics of objects in visual scenes in comparison to text}.
\newblock \emph{Artificial Intelligence}, 274:44--65.

\bibitem[{Miller(1995)}]{wordnet}
George~A Miller. 1995.
\newblock Wordnet: a lexical database for english.
\newblock \emph{Communications of the ACM}, 38(11):39--41.

\bibitem[{Mohammad(2018)}]{vad}
Saif Mohammad. 2018.
\newblock Obtaining reliable human ratings of valence, arousal, and dominance
  for 20,000 english words.
\newblock In \emph{Proceedings of the 56th Annual Meeting of the Association
  for Computational Linguistics (Volume 1: Long Papers)}, pages 174--184.

\bibitem[{Pezzelle et~al.(2021)Pezzelle, Takmaz, and
  Fernández}]{10.1162/tacl_a_00443}
Sandro Pezzelle, Ece Takmaz, and Raquel Fernández. 2021.
\newblock \href {https://doi.org/10.1162/tacl_a_00443} {{Word Representation
  Learning in Multimodal Pre-Trained Transformers: An Intrinsic Evaluation}}.
\newblock \emph{Transactions of the Association for Computational Linguistics},
  9:1563--1579.

\bibitem[{Radford et~al.(2021)Radford, Kim, Hallacy, Ramesh, Goh, Agarwal,
  Sastry, Askell, Mishkin, Clark et~al.}]{radford2021learning}
Alec Radford, Jong~Wook Kim, Chris Hallacy, Aditya Ramesh, Gabriel Goh,
  Sandhini Agarwal, Girish Sastry, Amanda Askell, Pamela Mishkin, Jack Clark,
  et~al. 2021.
\newblock Learning transferable visual models from natural language
  supervision.
\newblock In \emph{International Conference on Machine Learning}, pages
  8748--8763. PMLR.

\bibitem[{Radford et~al.(2019)Radford, Wu, Child, Luan, Amodei, and
  Sutskever}]{gpt}
Alec Radford, Jeff Wu, Rewon Child, David Luan, Dario Amodei, and Ilya
  Sutskever. 2019.
\newblock Language models are unsupervised multitask learners.

\bibitem[{Ruytenbeek et~al.(2017)Ruytenbeek, Verheyen, and
  Spector}]{ruytenbeek2017asymmetric}
Nicolas Ruytenbeek, Steven Verheyen, and Benjamin Spector. 2017.
\newblock Asymmetric inference towards the antonym: Experiments into the
  polarity and morphology of negated adjectives.
\newblock \emph{GLOSSA-A JOURNAL OF GENERAL LINGUISTICS}, 2(1).

\bibitem[{Speer et~al.(2017)Speer, Chin, and Havasi}]{conceptnet}
Robyn Speer, Joshua Chin, and Catherine Havasi. 2017.
\newblock Conceptnet 5.5: An open multilingual graph of general knowledge.
\newblock In \emph{Thirty-first AAAI conference on artificial intelligence}.

\bibitem[{Tan and Bansal(2019)}]{tan2019lxmert}
Hao Tan and Mohit Bansal. 2019.
\newblock Lxmert: Learning cross-modality encoder representations from
  transformers.
\newblock \emph{arXiv preprint arXiv:1908.07490}.

\bibitem[{Tan and Bansal(2020)}]{tan2020vokenization}
Hao Tan and Mohit Bansal. 2020.
\newblock Vokenization: Improving language understanding with contextualized,
  visual-grounded supervision.
\newblock \emph{arXiv preprint arXiv:2010.06775}.

\bibitem[{Terkourafi et~al.(2020)Terkourafi, Weissman, and
  Roy}]{terkourafi2020different}
Marina Terkourafi, Benjamin Weissman, and Joseph Roy. 2020.
\newblock Different scalar terms are affected by face differently.
\newblock \emph{International Review of Pragmatics}, 12(1):1--43.

\bibitem[{Van~Tiel and Pankratz(2021)}]{van2021adjectival}
Bob Van~Tiel and Elizabeth Pankratz. 2021.
\newblock Adjectival polarity and the processing of scalar inferences.
\newblock \emph{Glossa: a journal of general linguistics}, 6(1).

\bibitem[{Vuli{\'c} et~al.(2020)Vuli{\'c}, Ponti, Litschko, Glava{\v{s}}, and
  Korhonen}]{vulic2020probing}
Ivan Vuli{\'c}, Edoardo~Maria Ponti, Robert Litschko, Goran Glava{\v{s}}, and
  Anna Korhonen. 2020.
\newblock Probing pretrained language models for lexical semantics.
\newblock \emph{arXiv preprint arXiv:2010.05731}.

\end{thebibliography}
\bibliographystyle{acl_natbib}

\clearpage
\onecolumn
\appendix

\section{Properties of embedding spaces}

\centering
\includegraphics[height=.65\textheight]{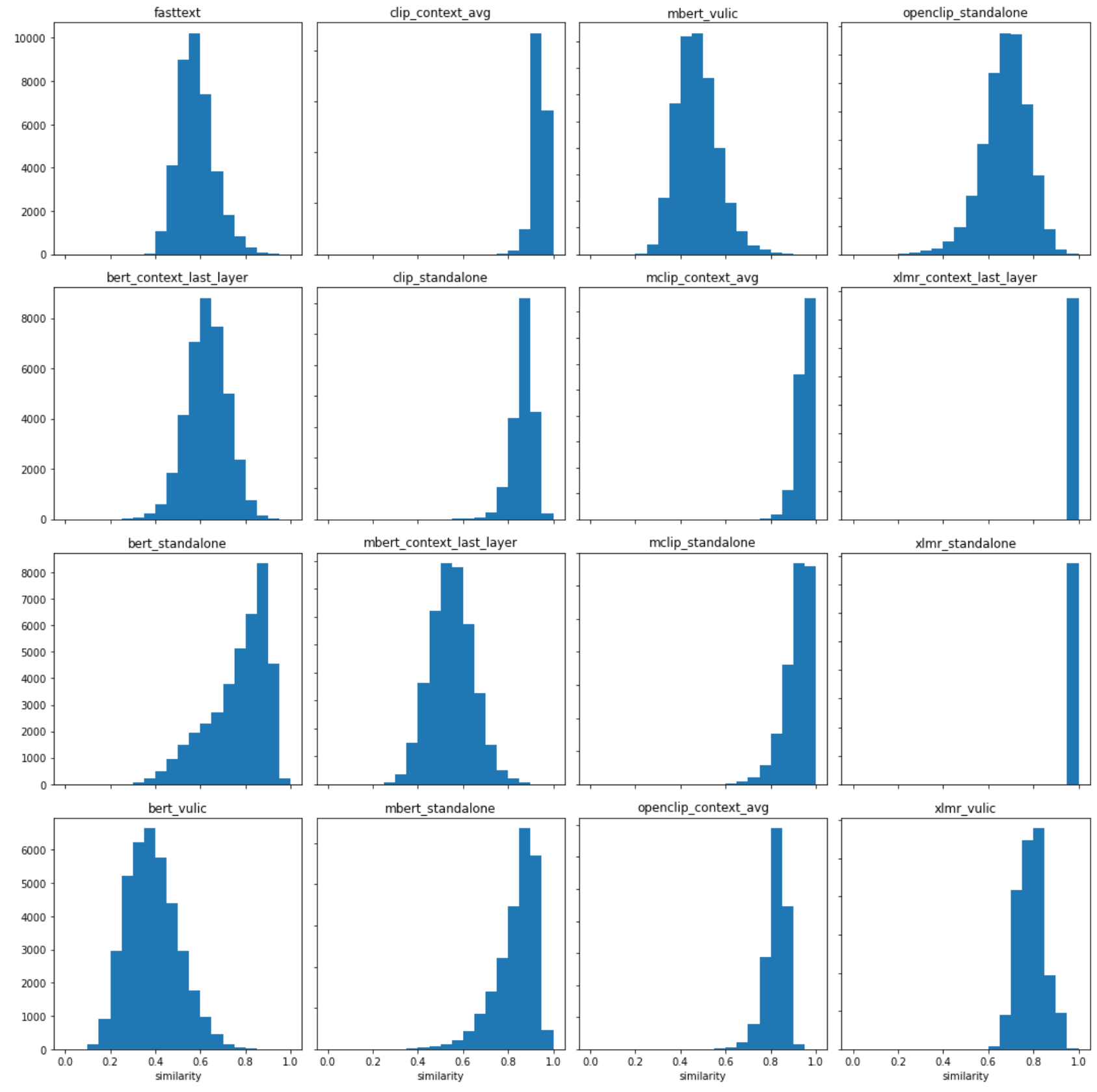}
Distributions of similarities between lexical pairs per model and embedding type

\section{Plots for more factors}
\includegraphics[width=.23\textwidth]{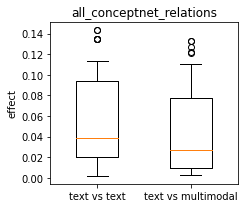} \includegraphics[width=.23\textwidth]{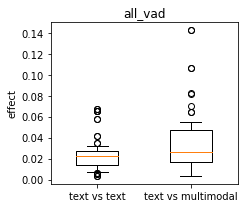}  
\includegraphics[width=.23\textwidth]{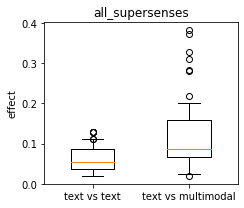} \includegraphics[width=.23\textwidth]{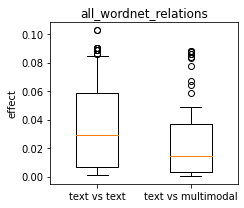}

Comparing semantic features' contributions to contrasts between text models vs.~ other text models, on the one hand, and text models vs. L\&V models, on the other hand. Explanatory contributions of ConceptNet relations, combined VAD features, combined WordNet supersenses, and combined WordNet relations.
\end{document}